%% file: main.tex
\documentclass[10pt,twocolumn,letterpaper]{article}

\usepackage[pagenumbers]{cvpr} 

\input{preamble}

\definecolor{cvprblue}{rgb}{0.21,0.49,0.74}
\usepackage{multirow}
\usepackage[ruled,linesnumbered]{algorithm2e}
\usepackage{array}
\usepackage[colorlinks,bookmarksopen,bookmarksnumbered,citecolor=green, linkcolor=red, urlcolor=green]{hyperref}

\def\confName{CVPR}
\def\confYear{2024}

\title{An Empirical Study of Scaling Law for OCR}

\author{%
	Miao Rang$^{*}$\quad Zhenni Bi\thanks{These authors contributed equally to this work.} \quad Chuanjian Liu \quad 
	Yunhe Wang$^{\dag}$\quad Kai Han\thanks{Corresponding Author.} \\
	Huawei Noah's Ark Lab\\
	\texttt{\{rangmiao1,bizhenni,liuchuanjian,} \\
		\texttt{yunhe.wang,kai.han\}@huawei.com} \\
}

\begin{document}
\maketitle
\input{sec/0_abstract}    
\input{sec/1_intro}

\input{sec/2_relatedwork}
\input{sec/3_methold}
\input{sec/4_result_analyse}
\input{sec/5_discussion}
\input{sec/X_suppl}

{
    \small
    \bibliographystyle{ieeenat_fullname}
    \bibliography{main}
}


\end{document}

%% file: preamble.tex
\usepackage[dvipsnames]{xcolor}


%% file: sec/0_abstract.tex
\begin{abstract}
The laws of model size, data volume, computation and model performance have been extensively studied in the field of Natural Language Processing (NLP). However, the scaling laws in Optical Character Recognition (OCR) have not yet been investigated. To address this, we conducted comprehensive studies that involved examining the correlations between performance and the scale of models, data volume and computation in the field of text recognition. Conclusively, the study demonstrates smooth power laws between performance and model size, as well as training data volume, when other influencing factors are held constant. Additionally, we have constructed a large-scale dataset called \textbf{REBU-Syn}, which comprises 6 million real samples and 18 million synthetic samples. Based on our scaling law and new dataset, we have successfully trained a scene text recognition model, achieving a new state-of-the-art on 6 common test benchmarks with a top-1 average accuracy of 97.42$\%$. The models and dataset are publicly available at \href{https://github.com/large-ocr-model/large-ocr-model.github.io}{large-ocr-model.github.io}.
\end{abstract}

%% file: sec/1_intro.tex
\section{Introduction}
\label{sec:intro}
Optical Character Recognition (OCR) is a technology designed to detect and interpret textual content within images, such as scanned documents and photographs. A typical OCR system operates in two primary phases: text detection and text recognition. In this context, our focus is on the text recognition phase, which involves identifying and extracting text from predefined bounding boxes. In text recognition tasks, it is subdivided into scene text recognition and scanned document recognition. Scene text recognition is more challenging than traditional scanned documents because it needs to solve more complex problems in practical applications, such as illumination changes, occlusion, distortion, angle changes and other factors. The impact makes the identification task more difficult. Moreover, STR, as an active research field, has greater room for technological improvement and innovation. Consequently, in order to investigate the applicability of the scaling law, this paper focuses on the scene text recognition task.


 
\begin{figure}[t]
  \centering
   \includegraphics[width=0.9\linewidth]{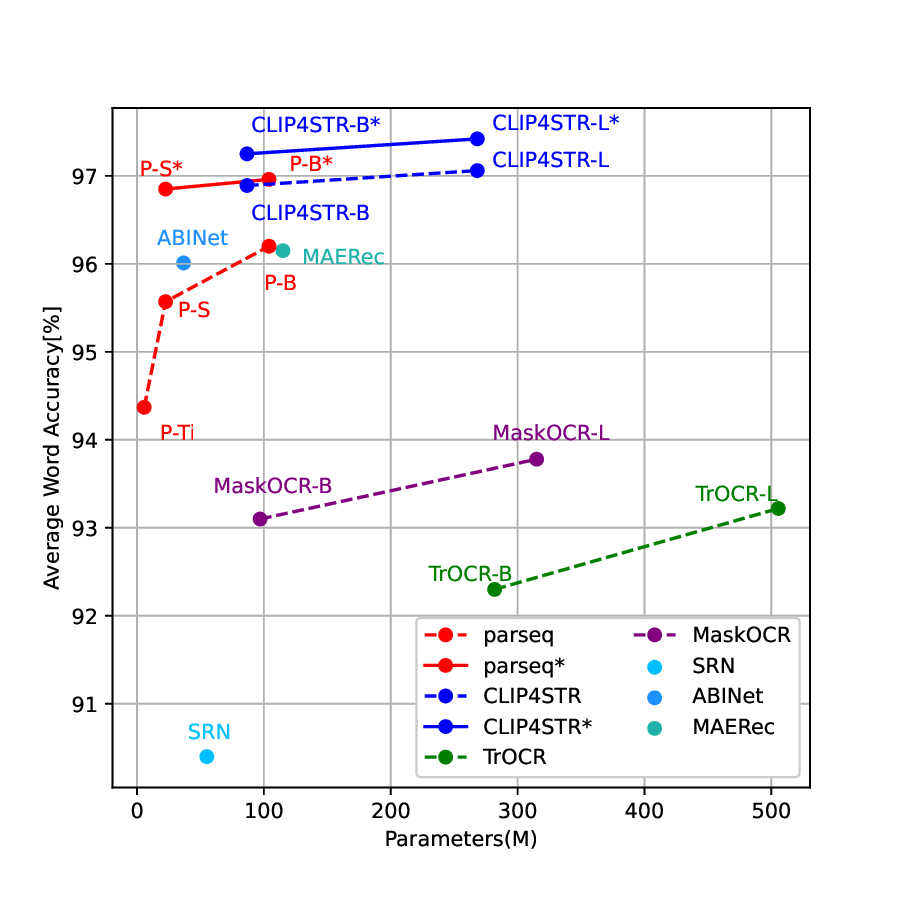}
    \setlength{\abovecaptionskip}{0cm}
   \caption{Mean word accuracy vs Parameters on the 6 common test benchmarks. P-Ti, P-S and P-B refer to PARSeq-Ti, PARSeq-S and PARSeq-B, respectively. * indicates training with REBU-Syn.}
   \label{fig:onecol}
   \vspace{-1em}
\end{figure}


With the introduction of large-scale models in deep learning, an increasing number of academics are focusing on the potential and growth trends of these models, hoping that they will contribute to the development and design of future models. In the field of NLP~\cite{bao2021beit, brown2020language},
 numerous experiments have been carried out to investigate scaling model laws~\cite{kaplan2020scaling, zhai2022scaling, dehghani2023scaling,ghorbani2021scaling}. The results show that the larger the volume of data fed into the neural network, the better its performance. Therefore, large language models trained on vast amounts of data have dominated the field of NLP. However, in the OCR domain, research predominantly focuses on enhancing model performance using fixed data sizes and model sizes~\cite{shi2016end, baek2021use, li2021trocr}. Studies specifically addressing the scaling laws in large OCR models are noticeably sparse, which casts uncertainty on the potential impact of large-scale models and substantial data volumes in OCR. Transformer-based models have achieved state-of-the-art performance in various text recognition tasks and challenges~\cite{li2021trocr, sheng2019nrtr, tan2022pure,raisi2021transformer}.

In this paper, we explore the scaling laws of text recognition using transformer-based models. Our focus is on unraveling the relationships between model size, data volume and computation in relation to model performance. Our experimental framework encompasses a wide range of models, with parameter counts ranging from 50 million to 1 billion, and datasets that vary from 1 million to 1 billion training samples. Additionally, we extend our exploration to computational durations ranging from 100 to 1000 hours. This comprehensive analysis allows us to draw insightful conclusions about the scaling law in text recognition. Furthermore, we introduce a novel dataset called \textbf{REBU-Syn}, which combines real-world and synthetic data. This dataset has been meticulously compiled from existing public datasets, providing a valuable resource for further research in this field.

Throughout our research, we develop an advanced method for large-scale training. This method involves a comprehensive examination of various strategies, such as optimizing training hyperparameters, analyzing data distributions and utilizing pre-training techniques. Our objective is to create a model characterized by exceptional precision and accuracy. The culmination of these efforts is the training of CLIP4STR-L using REBU-Syn. This approach results in achieving a groundbreaking state-of-the-art performance of 97.42$\%$ on the test benchmark (see Fig~\ref{fig:onecol}). The following is a compilation of additional scaling laws for OCR observations:
\begin{itemize}
    \item The scaling law holds in the field of OCR. There exist smooth power laws between the size of the model, the volume of data, computation and performance.
    \item Large-scale models make better use of samples than small-scale models which means that large models achieve lower error rate with fixed amount of data.
    \item The proportion of training data from different sources is crucial for model training.
    \item Models pre-trained on OCR-related data are more effective in OCR tasks than models pretrained on general images like ImageNet.
\end{itemize}

%% file: sec/2_relatedwork.tex
\section{Related Work}

\paragraph{Model Scale}
Recent research has extensively explored the scaling laws for Transformer language models, particularly in the field of NLP~\cite{liu2018generating,kaplan2020scaling}. These studies have established a set of universal principles for modeling scale. However, research specifically addressing OCR remains scarce. Transformer-based methods, known for their higher tolerance to increased model depth and width, have been applied in various fields~\cite{dehghani2018universal,child2019generating,devlin2018bert,raffel2020exploring,lepikhin2020gshard}. This study leverages these methods, with a specific focus on their application in OCR, to provide guidance on making effective adjustments in model size.

\paragraph{Data Scale}
In the domain of image recognition, the scale of data plays a critical role. The performance of various models is significantly influenced by the size of the datasets used~\cite{russakovsky2015imagenet, recht2019imagenet, barbu2019objectnet}. While different model types require varying data volumes, some previous methods~\cite{baek2019wrong} explored the impact of OCR recognition tasks on different data scales, but their main focus was on CNN-based or attention-based approaches~\cite{shi2016end,simonyan2014very, borisyuk2018rosetta,cheng2017focusing,lee2016recursive,liu2018char}, and they focus solely on reducing the data scale. Furthermore, the availability of public datasets has facilitated extensive research and experimentation in this field~\cite{russakovsky2015imagenet, recht2019imagenet, barbu2019objectnet}. This paper aims to build upon these foundations by conducting a comprehensive investigation into the effects of data scale, both at the lower and upper limits, as well as the distribution of data in OCR tasks. Additionally, this study offers new insights into the alignment of real and synthetic data during the training of optimal models, filling a gap in current research.

\paragraph{Scaling Laws}
The rapid advancement of Large Language Models (LLMs) like ChatGPT~\cite{chatgpt} and GPT-4~\cite{chatgpt4} has sparked research into universal scaling laws~\cite{kaplan2020scaling, lepikhin2020gshard} in deep learning. These studies explore the relationship between model size, data volume, computation and performance, providing training principles for large models in NLP. ~\cite{henighan2020scaling} describes laws for autoregressive generative modeling. Similar scaling theories have also emerged in the field of computer vision~\cite{zhai2022scaling}, as demonstrated by the training of ViT-G with 2B parameters~\cite{zhai2022scaling}. Furthermore, recent work has been done on the scaling law of CLIP~\cite{radford2021learning} has revealed task- and dataset-dependent scaling behaviors~\cite{li2023inverse}. Building upon these foundational insights, this study represents a unique exploration of scaling laws within the context of OCR. Specifically, it explores the allocation of parameters and the internal structure of the transformer model, with the aim of optimizing performance for text recognition. This investigation makes a unique contribution to the expanding body of research on scaling laws, particularly in the underexplored domain of OCR.

%% file: sec/3_methold.tex
\section{Method Details}
In this paper, our primary focus is to explore the scaling laws for the transfer performance of Transformer-based models in text recognition tasks. Concurrently, we have amalgamated all publicly available datasets to construct the REBU-Syn dataset. This paper also includes a thorough analysis of the data proportions obtained from various sources. Finally, we will provide a detailed overview of the training parameter settings used in our study.

\subsection{Model Scaling}
\paragraph{TrOCR}
TrOCR~\cite{li2021trocr} is a text recognition model that utilizes pure Transformer architecture. It integrates pre-trained Computer Vision (CV) and NLP models. And it is the first work that jointly leverages image Transformer and text Transformer for text recognition tasks in OCR. The scaling laws of Transformer language models~\cite{kaplan2020scaling} and Vision Transformers~\cite{zhai2022scaling} have been studied, the scaling laws for models in the OCR field have not yet been explored. Based on this, we scaled the TrOCR model size and attempted to analyze the accuracy change curves of models with varying sizes. 

In TrOCR, the encoder and decoder parts utilize pre-trained image Transformer and text Transformer models, respectively. These pre-trained models utilize large-scale unlabeled data for image understanding and language modeling. As a result, TrOCR does not require additional language models for post-processing, and the model outperforms current state-of-the-art models in tasks related to printing and handwriting recognition. In order to continue benefiting from pre-training for related tasks, we select the most suitable combination of encoder and decoder in TrOCR for scaling.

For TrOCR-S, we use DeiT$_{SMALL}$~\cite{touvron2021training} to initialize the encoder and MiniLM~\cite{wang2020minilm} to initialize the decoder. TrOCR-B uses BEIT$_{BASE}$~\cite{bao2021beit} to initialize the encoder and RoBERTa$_{LARGE}$~\cite{liu2019roberta} to initialize the decoder. TrOCR-L and TrOCR-H utilize BEIT$_{LARGE}$ to initialize the encoder and RoBERTa$_{LARGE}$ to initialize the decoder. The model's parameters range from 43.09 million to 1 billion, and the details of the parameters are shown in Table~\ref{trocr-model-setting-table}.

\begin{table}[ht!]
	\begin{center}
            \scalebox{0.7}{
    		\begin{tabular}{c|c|c|c|c|c}
    			\hline
    			\multirow{2}{*}{\bf{Model}}     &\multicolumn{3}{c|}{\bf{Encoder}} & \multirow{2}{*}{\bf{FLOPs (G)}} & \multirow{2}{*}{\bf{Params (M)}}  \\
    			\cline{2-4} 
    			& layers & hidden sizes & heads & \\
    			
    			\hline
    			TROCR-S & 12 & 384 & 6 & 13.31  & 43.09 \\
    			TROCR-B & 12 & 768 & 12 & 62.01 & 281.87 \\
    			TROCR-L & 24 & 1024 & 16 & 191.00      & 505.50 \\
    			TROCR-H & 48 & 1200 & 16 &  497.91       & 1037.61 \\
    			\hline
    		\end{tabular}
            }
	\end{center}
       \setlength{\abovecaptionskip}{-0.1cm}
       \setlength{\belowcaptionskip}{-0.65cm}
	\caption{Architecture specifications of TrOCR variants.}
	\label{trocr-model-setting-table}
\end{table}

\paragraph{PARSeq}
PARSeq~\cite{bautista2022scene} follows an encoder-decoder architecture. PARSeq is also based on the Transformer framework with excellent accuracy, which perfectly fits the scope of our research. The encoder part utilizes the Vision Transformer (ViT) model to extract image features, while the decoder follows the same architecture as the pre-LayerNorm~\cite{wang2019learning}. Transformer decoder in this study utilizes twice the number of attention heads, where nhead = dmodel/32. In contrast to the standard ViT, the encoder removes the $[class]$ token and inputs all the output tokens into the decoder.

PARSeq has two models in the original paper, PARSeq-Ti and PARSeq-S. In order to investigate the law of large models in the field of OCR, the scaling law of the ViT model was demonstrated. ~\cite{zhai2022scaling}. Based on this, we scaled PARSeq to 4 different sizes. On the basis of the original paper PARSeq-S, the model was expanded to 3 sizes: PARSeq-B, PARSeq-L, and PARSeq-H. The scale of the model was also expanded from 22 million to 0.6 billion. The configurations with different scale PARSeq models can be seen in Table~\ref{parseq-model-setting-table}.

\begin{table}[ht!]
	\begin{center}
            \scalebox{0.7}{
    		\begin{tabular}{c|c|c|c|c|c}
    			\hline
    			\multirow{2}{*}{\bf{Model}}  &\multicolumn{3}{c|}{\bf{Encoder}} & \multirow{2}{*}{\bf{FLOPs (G)}} & \multirow{2}{*}{\bf{Params (M)}}  \\
    			\cline{2-4} 
    			 & layers & hidden sizes & heads &  \\
    			
    			\hline
    			PARSeq-S & 12 & 384 & 6 & 2.76  & 22.51 \\
    			PARSeq-B & 12 & 768 & 12 & 17.20 & 104.01 \\
    			PARSeq-L & 24 & 1024 & 16 & 49.90 & 335.92 \\
    			PARSeq-H & 32 & 1280 & 16 & 98.10 & 682.14 \\
    			\hline 			
    		\end{tabular}
            }
	\end{center}
        \vspace{-10pt}
        \setlength{\abovecaptionskip}{0.2cm}
        \setlength{\belowcaptionskip}{-0.5cm}
	\caption{Architecture specifications of PARSeq variants.}
	\label{parseq-model-setting-table}
\end{table}

\subsection{Dataset}
\paragraph{Training Dataset}

The training datasets for text recognition are typically categorized into synthetic and real data. Historically, scene text recognition models primarily relied on synthetic data due to the scarcity of real-world data. However, the recent increase in the availability of real data has shifted this trend. It has been observed that models trained on real data tend to be more sample-efficient compared to those trained on synthetic data. In light of this, we meticulously collected both synthetic and real data, employing various strategies to construct the REBU-Syn dataset. This dataset comprises approximately 6M real data samples and 18M public synthetic data samples, as detailed in Table~\ref{dataset_volume-table}. The ratio of synthetic to real data in REBU-Syn is 3:1. Furthermore, we utilized synthesis technology to generate an additional 60M data samples, similar to MJST, termed $MJST^{+}$.

\textbf{Real Dataset} We gather real images from 4 widely-accessible datasets to assemble the \textbf{REBU}. The \textbf{R} component consists of commonly used real data~\cite{bautista2022scene}, including COCO-Text (COCO)~\cite{veit2016cocotext}, RCTW17 ~\cite{shi2018icdar2017}, Uber-Text (Uber)~\cite{zhang2017uber}, ArT~\cite{chng2019icdar2019}, LSVT~\cite{sun2020chinese}, MLT19~\cite{nayef2019icdar2019}, ReCTS~\cite{liu2019icdar}, TextOCR~\cite{singh2021textocr} and OpenVINO~\cite{krylov2021open}. A detailed analysis of these datasets is presented in ~\cite{bautista2022scene}. \textbf{U}, another segment of REBU, includes 4 million labeled images across 14 datasets, collectively referred to as Union14M-L~\cite{bautista2022scene}. \textbf{B} represents the training data from benchmark sources, encompassing datasets such as IIIT 5k-word (IIIT5k)~\cite{mishra2012scene}, Street View Text (SVT)~\cite{wang2011end}, ICDAR13~\cite{karatzas2013icdar} and ICDAR15~\cite{karatzas2015icdar}. Furthermore, \textbf{E} is composed of images from two commonly used real datasets in text detection tasks, namely Total Text~\cite{ch2017total} and CTW1500~\cite{yuliang2017detecting}.This inclusion significantly expands the range of real data in our collection.

\textbf{Public Synthetic Dataset} MJSynth (MJ)~\cite{jaderberg2014synthetic} and SynthText (ST)~\cite{gupta2016synthetic} are two widely-utilized synthetic datasets in the field of scene text recognition, containing 8.9M million and 5.5M million data samples respectively. Addtionly, we incorporated two other composite datasets into our study. Curved SyntheText (CST) and SyntheAdd (SA)~\cite{huang2022swintextspotter}. CST is specifically designed for text detection tasks, primarily comprising curved text data. SA, generated with the SynthText engine, is aimed at synthesizing less common characters, including punctuation marks.

\textbf{Generate Synthetic Dataset} To closely align with the MJ and ST datasets, we created $MJST^{+}$ using two data generation tools: TextRecognitionDataGenerator\footnote{https://github.com/Belval/TextRecognitionDataGenerator} and SynthText\footnote{https://github.com/ankush-me/SynthText}. The TextRecognitionDataGenerator is adept at producing data that mimics complex scenes, encompassing effects such as blurring, tilting and distortion. SynthText, on the other hand, specializes in synthesizing data akin to ST, resulting in samples that blend more seamlessly with natural scenes.

To augment the diversity of the generated corpus, we sourced 700,000 corpus entries from the most extensively utilized English corpus website globally\footnote{https://www.english-corpora.org/corpora.asp}. For the background selection in our synthesized images, we employed natural scene pictures provided by SynthText as the backdrop. Utilizing these two synthesis methods, we successfully synthesized a total of 60M data samples. Code for data synthesis is available\footnote{https://github.com/large-ocr-model/large-ocr-model.github.io}.

\begin{table}[ht!]
	\begin{center}
            \scalebox{0.7}{
    		\begin{tabular}{c|c|c}
    			\hline
    			{\bf{Source}} & {\bf{Dataset}} & {\bf{Instances}}  \\

    			\hline
    			Public Real & Real (R) & 3.3M \\
    			Public Real & Extra Real Data (E) & 15k \\
    			Public Real & BenchMark (B) & 7.5K \\
    			Public Real & Union14M (U) & 3.1M \\
    			\hline
    			Public Synthetic & MJ & 5.5M \\
    			Public Synthetic & ST & 8.9M \\
    			Public Synthetic & CST & 1.8M \\
    			Public Synthetic & SA & 1.2M \\
                    \hline
    		  Generate Synthetic & $MJST^{+}$ & 60M \\
                    \hline    			
    		\end{tabular}            
            }
	\end{center}
        \vspace{-10pt}
        \setlength{\belowcaptionskip}{-0.8cm}
	\caption{Statistics of REBU-Syn datasets, including Public Real and Public Synthetic. Generate Synthetic can be used additionally.}
	\label{dataset_volume-table}
\end{table}

\paragraph{Test Dataset}
To assess the performance of our model, we utilized 6 publicly available real scene text datasets: IIIT5k-Words (IIIT5k)~\cite{mishra2012scene}, Street View Text (SVT)~\cite{wang2011end}, ICDAR 2013 (IC13)~\cite{karatzas2013icdar}, ICDAR 2015 (IC15)~\cite{karatzas2015icdar} , SVT-Perspective (SVTP)~\cite{phan2013recognizing} and CUTE80 (CUTE)~\cite{risnumawan2014robust}. Both the IC13 and IC15 test sets have various subdivisions. We follow the division proposed by Yu \emph{et al}~\cite{yu2020towards}, using a version of the IC15 test set containing 1,811 images, and the IC13 test set comprising 857 images.

However, to address challenges posed by differing annotation formats and the presence of duplicate, non-Latin, and damaged samples, we employed the following data fusion strategy:
\begin{itemize}
 \item \textbf{Polygonal Text} We sourced synthesized data from datasets used in text detection tasks with polygonal annotation boxes, such as Curved SyntheText, SyntheAdd and STR Benchmark.To adapt these polygonal texts for use, we improved upon the method proposed in~\cite{bautista2022scene}. Our approach involves identifying the minimum bounding box of the polygon and applying a perspective transformation, avoiding direct clipping using maximum and minimum coordinates. This method retains challenging samples, as suggested in~\cite{bautista2022scene}, while minimizing background interference, thus enabling the recognizer to focus on pertinent areas.

 \item \textbf{Remove invalid chars and samples} Focusing on Latin characters, which have extensive data availability, we retained samples composed only of letters and symbols. Samples not in our predefined dictionary were discarded.
\item \textbf{Remove duplicate data} As we integrated multiple datasets, some of which overlapped, we meticulously removed any duplicate entries.
\end{itemize}

\subsection{Experiment Settings}
We utilized the publicly available implementations of TrOCR and PARSeq as baseline models. To achieve optimal performance, we tailored the number of training epochs and adjusted the learning rates. The specific implementation details are as follows:

\textbf{Hyper-Parameters  }
For our experiments, we use V100 GPUs equipped with 32GB of memory to train all models. The learning rates are set differently for various models. Specifically, TrOCR-S is trained with a batch size of 1024 and a learning rate of 4e$-$4. TrOCR-B employs a batch size of 256 with a learning rate of 1e$-$4, and TrOCR-L operates with a batch size of 128 and a learning rate of 4$e$-5. We use BPE~\cite{sennrich2016neural} of Fairseq and SentencePiece~\cite{kudo2018sentencepiece} for tokenizing text lines into word pieces. For PARSeq models, a consistent learning rate of 7e$-$4 is used, with the batch size adjusted to be as close to 1024 as possible. 

\textbf{Evaluation Metrics}
Word accuracy was the primary metric for evaluating the datasets of scene text. In this work, we standardized the final output string to match the commonly used 36-character set (lowercase alphanumeric) to ensure a fair comparison across different models and datasets.

%% file: sec/4_result_analyse.tex
\section{Results and Analysis}
\subsection{Smooth Power Laws}

Model performance is primarily influenced by three variables: the number of model parameters $N$, the volume of the training data $D$, and the computation of the model $C$. In this section, we explore the power laws among these influential factors with model performance $E$. To effectively characterize the scaling of OCR models, we have conducted training with a variety of models, including TrOCR and PARSeq.

\begin{table}[ht!]
	
	\begin{center}
            \scalebox{0.7}{
    		\begin{tabular}{c|c|c|c|c|c|c|c}
    			\hline
    			 &  &\multicolumn{3}{c|}{\bf{Regular Text}}     &\multicolumn{3}{c}{\bf{Irregular Text}} \ \\
    			\cline{3-8}
    			{\bf{Model}}& {\bf{Avg}} &  IC13 & IIIT5k & SVT  & CUTE80 & IC15 & SVTP \\ 
    			
    			& & 857  & 3,000 & 647  & 288    & 1,811 & 645 \\
    			\hline
    			TrOCR-S & 81.93 & 90.65 & 85.60 & 85.94 & 74.31 & 72.73 & 78.44 \\
    			TrOCR-B & 88.56 & 96.14 & 92.00 & 91.56 & 80.56 & 81.14 & 83.91 \\
    			TrOCR-L & 89.84 & 96.50 & 92.90 & 92.81 & 84.38 & 82.52 & 86.72 \\
    			TrOCR-H & \bf{90.94} & 97.31 & 93.57 & 94.22 & 87.50 & 83.79 & 88.59 \\
    			\hline
    		\end{tabular}
            }
	\end{center}
       \setlength{\abovecaptionskip}{-0.1cm}
       \setlength{\belowcaptionskip}{-0.65cm}
	\caption{Word accuracy with different TrOCR model sizes. \emph{Train data}: Synthetic datasets with MJ and ST.}
	\label{trocr-model-table}
\end{table}
\begin{figure*}
  \centering
  \begin{subfigure}{0.33\linewidth}
      \includegraphics[width=0.99\linewidth] 
            {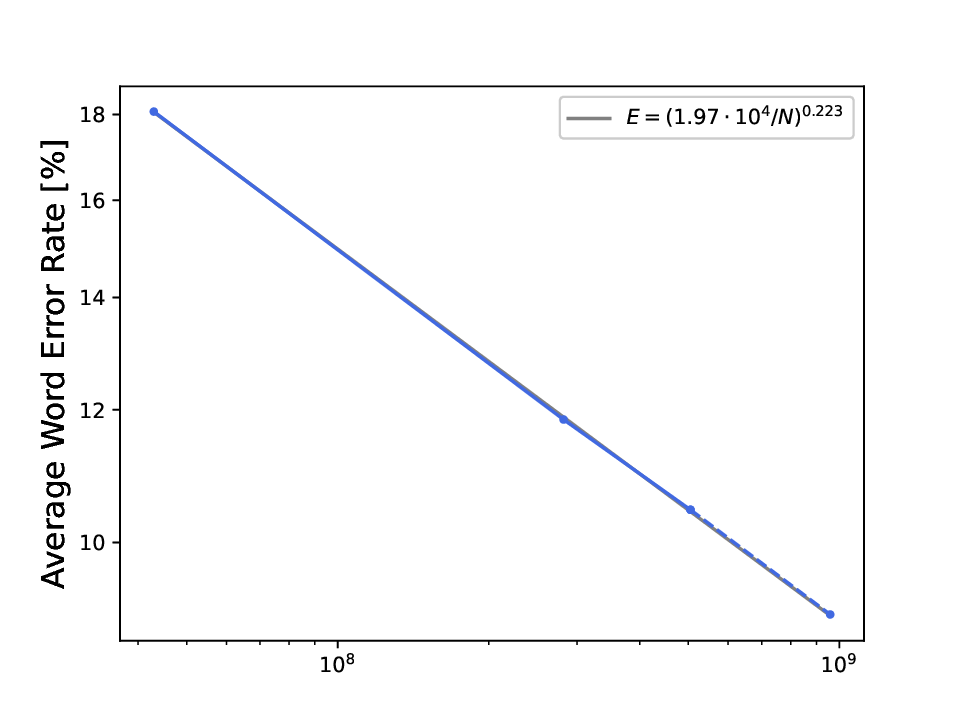} 
    \caption{Model size (Params)}
    \label{fig:short-a}
  \end{subfigure}
  \begin{subfigure}{0.33\linewidth}
      \includegraphics[width=0.99\linewidth] 
            {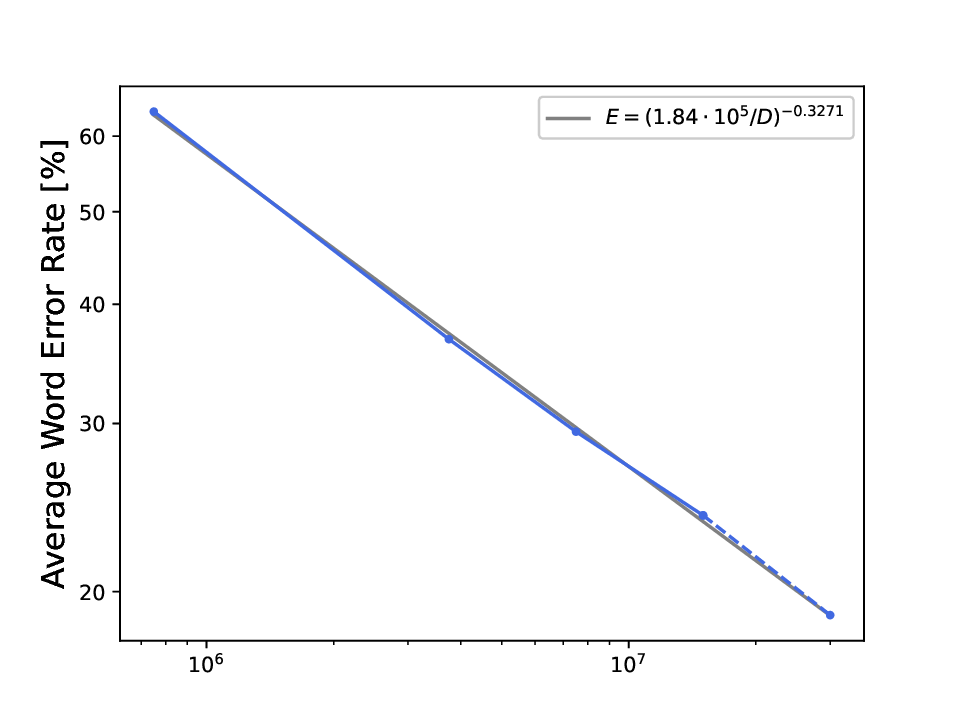} 
    \caption{Data volume (M)}
    \label{fig:short-b}
  \end{subfigure}
  \begin{subfigure}{0.33\linewidth}
      \includegraphics[width=0.99\linewidth] 
            {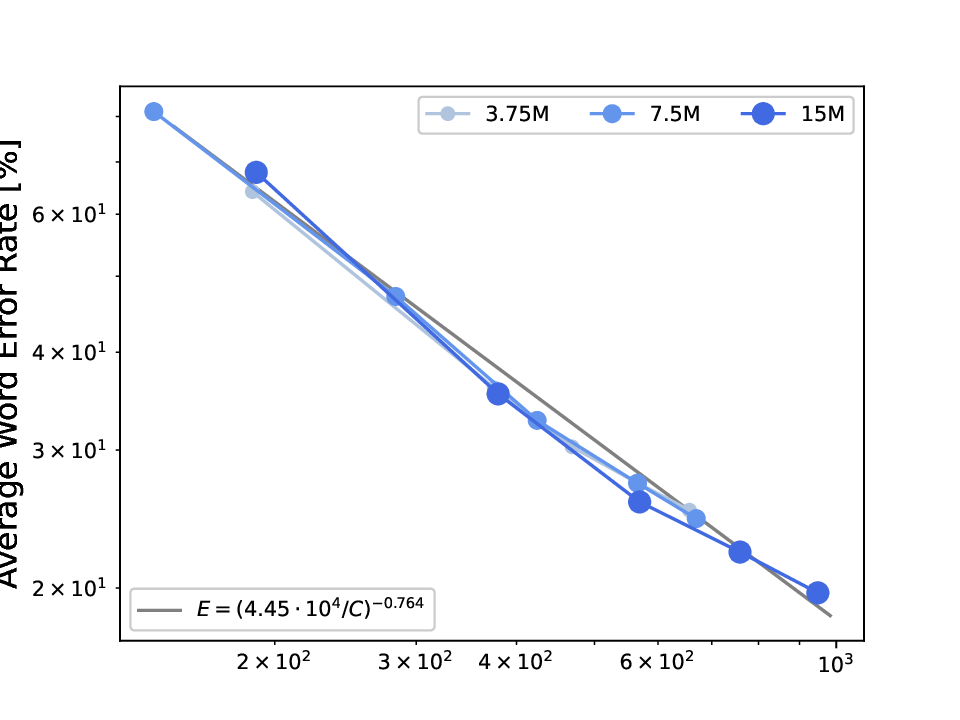} 
    \caption{Computation (training hours)}
    \label{fig:short-c}
  \end{subfigure}
  \hfill
 \setlength{\abovecaptionskip}{0cm}
  \caption{Improvement in TrOCR model performance with increasing model size, data volume, and training computation.  Model performance is measured by calculating the average word error rate on 6 common test benchmarks \textbf{Left}: Evaluation of model performance with changing model sizes.  
\textbf{Center}: Evaluation of model performance with varying data volumes. \textbf{Right}: Performance variations with different data sizes under varying computational resources. The x-axis represents the model's training time, measured in 8 GPU hours. For optimal performance, all three factors must be scaled up in tandem. Empirical performance exhibits a power-law relationship with each individual factor when it is not constrained by the other two factors.}
  \label{fig:short}
\end{figure*}

	
      

        


       
        
        


\DecMargin{1em} 

\subsubsection{The power law of model when data is fixed.}

\begin{itemize}
\item \textbf{Scaling TrOCR Models}
We trained 4 different scales (ranging in size from 43.09M to 1B) of TrOCR models. In order to maintain fairness and consistency with the experimental setting in the original TrOCR paper, we use MJ and ST to train TrOCR models with different model sizes. The experimental results on 6 common test benchmarks are shown in Table~\ref{trocr-model-table}. As shown in Fig~\ref{fig:short-a}, our analysis reveals a linear relationship on the log-log plot between the parameter count $N$ and modeling performance. This relationship can be described by a power-law equation ($E = aC^b$). Employing Algorithm~\ref{algo_disjdecomp} in the appendix, we utilized the first three models (TrOCR-S, TrOCR-B and TrOCR-L) to obtain the power function equation $E(\cdot)$. The last model (TrOCR-H) accurately aligns with the fitted straight line, demonstrating the effectiveness of the power law. The power law of the TrOCR model is as follows.
{\setlength\abovedisplayskip{2pt}
\begin{gather}
\label{eq1}
  	 E(N)=\left(1.97*10^{4}/N \right)^{0.223}  
\end{gather}
}
\end{itemize}

\begin{itemize}
\item \textbf{Scaling PARSeq Models}
To further validate the power law in relation to model parameters, we trained PARSeq models across 4 different scales with sizes ranging from 22M to 0.6B parameters, using the \textbf{REBU-Syn} dataset. The results of these experiments on 6 common test benchmarks are detailed in Table ~\ref{parseq-model-table}. As shown in Fig~\ref{fig:parseq_model}, the PARSeq demonstrates a similar trend to that observed with TrOCR, reinforcing the existence of the power law on model size. The power law of the PARSeq model is as follows.

 \begin{figure}[t]
  \centering
   \includegraphics[width=0.8\linewidth]{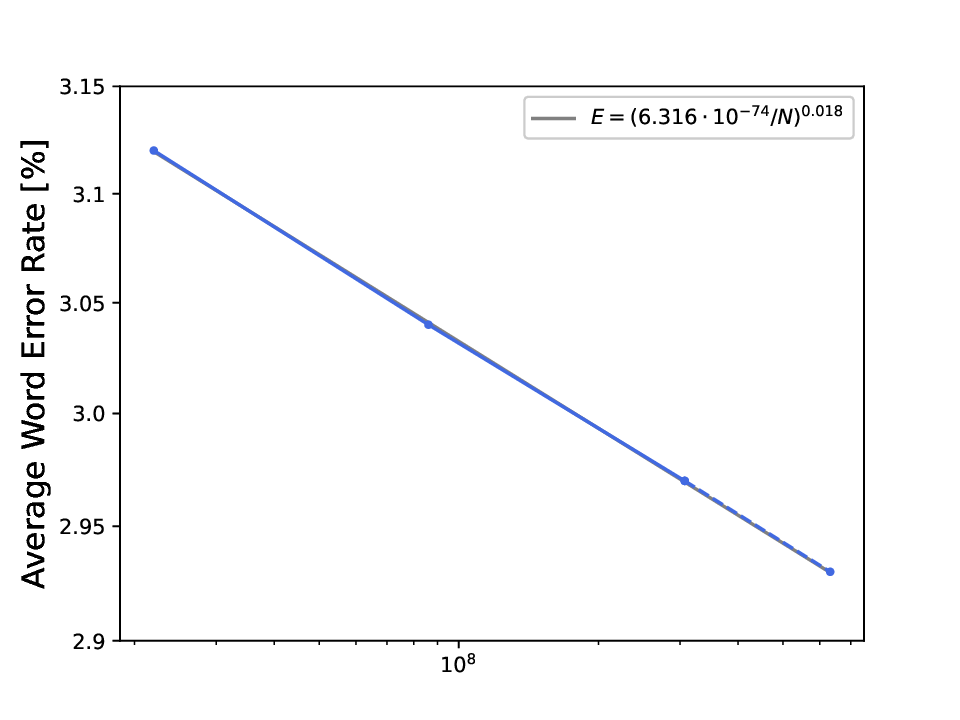}
    \setlength{\abovecaptionskip}{0cm}
    \setlength{\belowcaptionskip}{-0.3cm}
   \caption{The average word error rate on 6 common test benchmarks was calculated using the PARSeq model size. The solid line represents the fitted power law $E(\cdot)$, and the points on the dotted line correspond to the power law equation.}
   \label{fig:parseq_model}
\end{figure}

{\setlength\abovedisplayskip{0pt}
\setlength\belowdisplayskip{3pt}
\begin{gather}
\label{eq2}
  	 E(N)=\left(6.316*10^{-74}/N \right)^{0.018}  
\end{gather}
}
\end{itemize}

\begin{table}[ht!]
	\begin{center}
            \scalebox{0.7}{
    		\begin{tabular}{c|c|c|c|c|c|c|c}
    			\hline
    			 &   &\multicolumn{3}{c|}{\bf{Regular Text}}     &\multicolumn{3}{c}{\bf{Irregular Text}} \ \\
    			\cline{3-8}
    			{\bf{Model}}&{\bf{Avg}} &  IC13 & IIIT5k & SVT  & CUTE80 & IC15 & SVTP \\ 
    			
    			& & 857  & 3,000 & 647  & 288    & 1,811 & 645 \\
    			\hline
    			PARSeq-S & 96.85 & 98.72 & 98.63 & 98.45 & 99.65 & 92.27 & 95.97 \\
    			PARSeq-B & 96.96 & 99.07 & 98.53 & 98.76 & 99.31  & 92.44 & 96.74 \\
    			PARSeq-L & 97.03 & 99.30 & 98.63 & 98.61 & 99.31 & 92.32 & 97.21 \\
    			PARSeq-H & \bf{97.06} & 99.11 & 98.93 & 98.45 & 99.65 & 91.66 & 97.67 \\
    			\hline			
    		\end{tabular}
            }
	\end{center}
       \setlength{\abovecaptionskip}{-0.1cm}
       \setlength{\belowcaptionskip}{-0.5cm}
	\caption{Word accuracy with different model size of PARSeq. \emph{Train data}: REBU-Syn.}
	\label{parseq-model-table}
\end{table}

\subsubsection{The power-law of data when model size is fixed.}
\begin{itemize}
\item \textbf{Scaling Data Volume on TrOCR}
In order to explore the impact of data volume on model performance. We use MJ+ST and $MJST^{+}$ for training TrOCR-B. We randomly sampled data at various scales, with sizes 
ranging from 0.75M to 75M. The experimental results of TrOCR-B based on data of different scales are compiled in Table ~\ref{trocr-small-data-size-table}. We used various levels of data volume (solid blue line) to fit the power function (Eq.~\ref{eq3}) as shown by the solid grey line in ~\cref{fig:short-b}. The remaining portion of the data size (represented by the dashed blue line) still closely follows the power function, providing further evidence that the data volume adheres to the power function.

{\setlength\abovedisplayskip{0pt}
\setlength\belowdisplayskip{3pt}
\begin{gather}
\label{eq3}
  	 E(D)=\left(1.84*10^{5}/D \right)^{-0.3271}  
\end{gather}
}
\end{itemize}
\begin{table}[ht!]
	\begin{center}
            \scalebox{0.6}{
    		\begin{tabular}{c|c|c|c|c|c|c|c|c}
    			\hline
    			  &  &  &\multicolumn{3}{c|}{\bf{Regular Text}}     &\multicolumn{3}{c}{\bf{Irregular Text}} \ \\
    			\cline{4-9}
    			{\bf{Data Volume}}&\bf{Volume} &\bf{Avg}& IC13 & IIIT5k & SVT  & CUTE80 & IC15 & SVTP \\ 
    			
    			& && 857  & 3,000 & 647  & 288    & 1,811 & 645 \\
    			\hline
    			5$\%$ & 0.75M &50.35 & 64.53 & 57.77 & 47.60 & 29.51 & 39.98 & 38.14  \\
    			10$\%$ & 1.50M &52.61 & 64.18 & 59.43 & 55.63 & 33.68 & 42.13 & 42.33 \\
    			25$\%$ & 3.75M &74.86 & 86.11 & 79.30 & 80.22 & 60.07 & 63.83 & 71.47\\
    			50$\%$ & 7.50M &76.47 & 86.35 & 79.73 & 80.83 & 60.07 & 64.88 & 72.40 \\
    			100$\%$ & 15.00M &88.56 & 96.14 & 92.00 & 91.56 & 80.56 & 81.14 & 83.91 \\
    			500$\%$ & 75.00M & \bf{93.09} & 97.32 & 93.51 & 96.33 & 89.93 & 86.47 & 91.47 \\
    			\hline	
    		\end{tabular}
            }

	\end{center}
	\vspace{-10pt}
	\caption{TrOCR-B average accuracy in different percent of training data.}
	\label{trocr-small-data-size-table}
\end{table}

\begin{itemize}
\item \textbf{Scaling Data Volume on PARSeq}
Based on the power law of data volume, we utilize REBU-Syn in ParSeq-S training. By gradually expanding the data samples, the accuracy of PARSeq-S has been significantly improved in the Table ~\ref{parseq-small-data-size-table}.
\end{itemize}
\begin{table}[ht!]
	\begin{center}
            \scalebox{0.55}{
    		\begin{tabular}{c|c|c|c|c|c|c|c|c}
    			\hline
    			 &  &  &\multicolumn{3}{c|}{\bf{Regular Text}}     &\multicolumn{3}{c}{\bf{Irregular Text}} \ \\
    			\cline{4-9}
    			{\bf{Data}}& {\bf{Volume}}&{\bf{Avg}}& IC13 & IIIT5k & SVT  & CUTE80 & IC15 & SVTP \\ 
    			
    			& & & 857  & 3,000 & 647  & 288    & 1,811 & 645 \\
    			\hline
    			R & 3.30M & 95.57 & 97.32 & 97.87 & 97.37 & 97.22 & 90.34 &  94.73 \\
    			R+B+E & 3.32M & 95.63 & 97.43 & 97.97 & 97.84 & 98.96 & 90.28 & 93.64 \\
    			R+B+E+U & 6.42M & 96.12 & 99.53 & 97.93 & 97.53 & 98.96 & 91.39 & 95.66 \\
    			R+B+E+U+MJST & 20.82M & 96.45 & 98.48 & 98.40 & 97.84 & 98.61 & 91.44 & 96.43 \\
    			R+B+E+U+MJST+Syn & 23.82M & \bf{96.85} & 98.93 & 98.63 & 98.61  & 99.31 & 92.32 & 97.21 \\
    			\hline	
    		\end{tabular}         
            }
	\end{center}
         \setlength{\abovecaptionskip}{-0.1cm}
       \setlength{\belowcaptionskip}{-0.65cm}
	\caption{PARSeq-S average accuracy in different percent of training data.}
	\label{parseq-small-data-size-table}
\end{table}

\subsubsection{The power law of computation}
With the power laws of model size and data volume separately, we infer that the error rate and the compute budget can also be fit with the power law. We perform the study on TrOCR model. 
The outcome is depicted as the gray line on the plot on the right-hand side of~\cref{fig:short-c}. It can be fitted with the power formula as in Eq.~\ref{eq4}.

{\setlength\abovedisplayskip{0pt}
\setlength\belowdisplayskip{3pt}
\begin{gather}
\label{eq4}
    	E(C)=\left(4.45*10^{4}/ C \right)^{-0.3271} 
\end{gather}
}

\subsection{Other Observations}
\begin{figure}[t]
  \centering
   \includegraphics[width=0.9\linewidth]{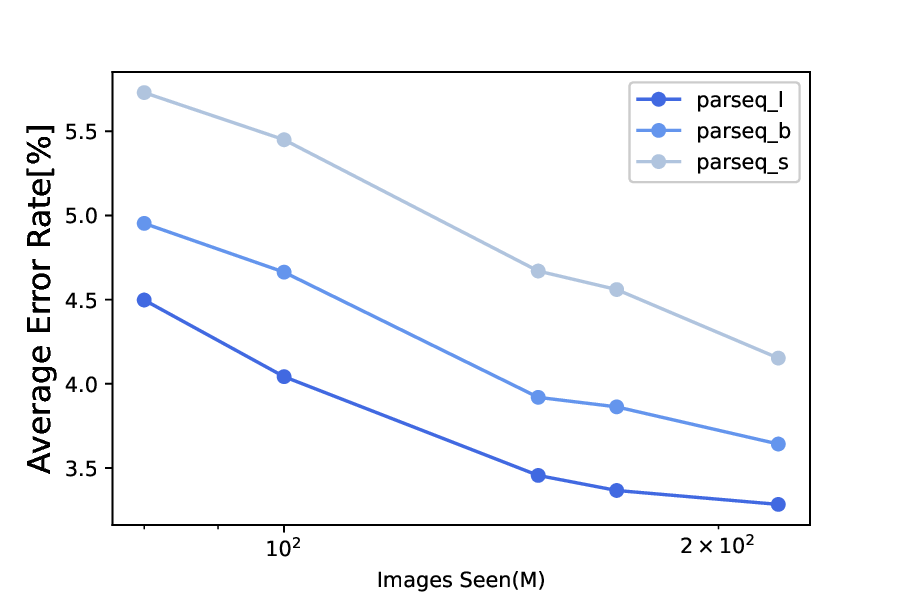}
    \setlength{\belowcaptionskip}{-0.2cm}
   \caption{Average word error rate on 6 common test benchmarks, with respect to images seen (batch size times number of steps) during PARSeq training stage of different sizes.}
   \label{fig:image_seen}
\end{figure}

\paragraph{Large-scale models make better use of samples.} As we continue to enlarge the model size, the model accuracy improves consistently. The phenomenon can be observed in Table~\ref{trocr-model-table} and ~\ref{parseq-model-table}. To improve training results, we can modify recognition models built on the Transformer architecture by utilizing the scaling laws of the vision Transformer. As shown in Figure~\ref{fig:image_seen} with respect to
the total number of images “seen” during PARSeq training stage of different sizes, it is clear that larger models utilize samples more effectively than their smaller models. When PARSeq models of different sizes are trained with an equal number of samples, smaller models exhibit a higher error rate compared to larger models. Furthermore, we observed that larger models tend to require fewer epochs to converge. For instance, PARSeq-S reached optimal accuracy in 32 epochs, whereas PARSeq-B needed only 14 epochs, and PARSeq-L just 5 epochs.  These findings suggest that with adequate training resources, it is more beneficial to train a larger model for fewer steps. This mirrors similar findings in language modelling~\cite{kaplan2020scaling} and machine translation~\cite{fernandes2023scaling}. However, when training time is a limiting factor, opting for a smaller model may be more practical.

\paragraph{The proportion of training data from various sources is crucial for model training.} The REBU-Syn comprises both real and synthetic data.  According to prior studies~\cite{bautista2022scene, zhao2023clip4str}, real data typically outperforms synthetic data in training efficiency, though synthetic data still plays a valuable role. Due to the high costs associated with obtaining and labeling real data, which often do not meet the volume requirements for model training, reliance on synthetic data is necessary. However, the effectiveness of synthetic data raises a question: Does more synthetic data always equate to better performance? Our findings suggest that an optimal ratio between real and synthetic data is crucial for enhanced model performance.

To achieve this objective, we conducted an experiment to investigate the proportional relationship between data obtained from various sources and determine the most efficient utilization of synthetic data. Synthetic data can be primarily categorized into two types: MJ+ST and Syn (CST+SA). MJ+ST, characterized by its large volume but homogeneous nature (consisting mostly of straight and clear samples), contrasts with SynText, which has a smaller volume (only one-fifth of MJ+ST) and primarily consists of curved texts. To evaluate the impact of different synthetic data sources on model accuracy, we trained PARSeq using a combination of real data and these synthetic datasets. The results, as presented in Table~\ref{data-source-table}, are revealing. The accuracy achieved using real data combined with MJ+ST is 96.24$\%$, only marginally higher by 0.05$\%$ compared to using real data with Syn. Given that the volume of MJ+ST is five times that of Syn, this implies that complex synthetic data is more sample-efficient. By simultaneously utilizing synthetic data from both MJ+ST and SynText along with real data, we observed a substantial enhancement in the accuracy of PARSeq, elevating it to state-of-the-art levels. This combination of diverse synthetic data styles, when integrated with real data, expands the range of the training data distribution. Such comprehensive coverage effectively enhances the overall quality and performance of the model.

\begin{table}[ht!]
	\begin{center}
            \scalebox{0.7}{
    		\begin{tabular}{c|c|c|c}
    			\hline
    			{\bf{Real DataSet}} & {\bf{Syn DataSet}} & {\bf{Data Ratio}} & {\bf{Word Acc}} \\

    			\hline
    			R+E+B+U & Syn & 1:0.5 & 96.19 \\
    			R+E+B+U & MJ+ST & 1:2.5 & 96.24 \\
    			R+E+B+U & MJ+ST+Syn & 1:3 & \textbf{96.85} \\
    			\hline
    		\end{tabular}
            }
	\end{center}
        \vspace{-10pt}
        \setlength{\belowcaptionskip}{-0.1cm}
	\caption{PARSeq-S average accuracy of integrating diverse synthetic and real data types.}
	\label{data-source-table}
\end{table}

Additionally, we investigated the effect of different synthetic-to-real data ratios on the accuracy of PARSeq-S. We maintained a constant amount of real data and progressively increased the volume of synthetic data. The ratio of synthetic to real data varied from 0.5 to 5 times. These varying proportions were achieved through random sampling. To augment the total volume of synthetic data, we randomly selected 18M samples from $MJST^{+}$ and combined them with the synthetic data in REBU-Syn, culminating in a total of 36M synthetic data samples.

\begin{table}[ht!]
	\begin{center}
            \scalebox{0.8}{
    		\begin{tabular}{c|c}
    			\hline
    			{\bf{Data Ratio}} & {\bf{Word Acc}} \\

    			\hline
    			Real:Syn=1:0.5 & 96.32 \\
    			Real:Syn=1:1 &96.50 \\
    			Real:Syn=1:2 & 96.59 \\
    			Real:Syn=1:3 & \textbf{96.85} \\
    			Real:Syn=1:4 & 96.76 \\
    			Real:Syn=1:5 & 95.70 \\
    			\hline
    		\end{tabular}
            }
	\end{center}
        \vspace{-15pt}
	\caption{PARSeq-S average accuracy on 6 common test benchmarks with varying ratios of synthetic and real data.}
	\label{data-ratio-table}
\end{table}

While synthetic data proves effective, it requires careful balancing with real data. As shown in Table~\ref{data-ratio-table}, a gradual increase in synthetic data led to some improvement in accuracy. Notably, the highest accuracy of 96.85$\%$ was achieved with a synthetic-to-real data ratio of 1:3. Beyond this ratio, the accuracy began to decline, likely due to the data distribution becoming overly skewed towards synthetic data, which can adversely affect model performance. \underline{Therefore, we recommend a synthetic-to-real data ratio of 1:3.} This balance offers the most significant improvement in accuracy without incurring excessive training costs.

\begin{table}[ht!]
	\begin{center}
            \scalebox{0.7}{
    		\begin{tabular}{c|c|c|c}
    			\hline
    			{\bf{Pretrain}} & {\bf{Dataset}}& {\bf{Backbone}}& {\bf{Word Acc}} \\

    			\hline
    			Scratch & R+E+B+U & ViT-S & 96.12 \\
   			Scratch & R+E+B+U+Syn & ViT-S & 96.85 \\
    			R+E+B+U+Syn & R+E+B+U & ViT-S & \textbf{96.96} \\
                    \hline
    			Scratch & R+E+B+U+Syn & ViT-L & 97.03 \\
    			ImageNet-21k & R+E+B+U+Syn & ViT-L & 96.74 \\
    			\hline
    		\end{tabular}
            }
	\end{center}
        \vspace{-10pt}
        \setlength{\belowcaptionskip}{-0.5cm}
	\caption{Average accuracy achieved by using visual task pre-training and OCR task pre-training on 6 common test benchmarks.}
	\label{pretrain-table}
\end{table}

\begin{table*}[ht]
	\begin{center}
            \scalebox{1}{
    		\begin{tabular}{c|c|c|c|c|c|c|c|c}
    			\hline
    			   &&\multicolumn{3}{c|}{\bf{Regular Text}}     &\multicolumn{3}{c|}{\bf{Irregular Text}} & \ \\
    			\cline{3-8}
    			{\bf{Method}} & {\bf{Training data}}  &  IC13 & IIIT5k & SVT  & CUTE80 & IC15 & SVTP & {\bf{Avg}} \\ 	
    			& &857  & 3,000 & 647  & 288    & 1,811 & 645&  \\
    			\hline
                    ViTSTR-B~\cite{atienza2021vision}& MJ+ST  & 92.40 & 88.40 & 87.70 & 81.30 & 72.60 & 81.80 & 85.46\\
                    SE-ASTER~\cite{qiao2020seed}& MJ+ST  & 92.80 & 93.80 & 89.60 & 83.60 & 80.0 &81.40 & 88.35\\
                    TRBA~\cite{baek2021use} &MJ+ST  & 93.4 & 92.10 & 88.90 & 89.20 & 77.4 & 89.30 & 90.07\\
                    SRN~\cite{SRN2020Yu} & MJ+ST  & 95.5 & 94.80 & 91.50 & 87.80 & 82.70 & 85.10 & 90.42 \\
                    TextScanner~\cite{TextScanner2019wan}& MJ+ST  & 94.90 & 95.70 & 92.70 & 91.60 & 83.50 & 84.80& 91.15 \\


                    VisionLAN~\cite{wang2021one}  & MJ+ST  & 95.7 & 95.80 & 91.70 & 88.50 & 83.7 & 86.00& 91.23 \\
                    PETR~\cite{wang2022petr} &MJ+ST  & 97.00 & 95.80 & 92.40 & 89.90 & 83.30 & 86.20& 91.42 \\
                    ABINet~\cite{fang2021read}& MJ+ST  & 97.4 & 96.20 & 93.50 & 89.20 & 86.00 & 89.30& 92.66 \\
                    MaskOCR(ViT-B)~\cite{lyu2022maskocr} & MJ+ST  & 98.1 & 95.8 & 94.7 & 89.2 & 87.3 & 89.9& 93.1 \\
                    PARSeq$_{A}$ ~\cite{bautista2022scene} & MJ+ST & 96.20 & 97.00 & 93.60 & 92.20 & 82.90 & 88.90 & 93.16 \\
                    MATRN~\cite{na2022multimodal} & MJ+ST  & 95.80 & 96.60 & 95.00 & 93.50 & 82.80 & 90.60 & 93.20\\
                    TrOCR$_{Large}$~\cite{li2021trocr} & MJ+ST+B  & 97.30 & 94.10 & 96.10 & 95.10 & 84.10 & 93.00& 93.23 \\


                    DiG-ViT-B~\cite{yang2023reading} & MJ+ST & 96.90 & 96.70 & 94.60 & 91.30 & 87.10 & 91.00 & 93.41 \\
                    MaskOCR(ViT-L)~\cite{lyu2022maskocr} & MJ+ST  & 98.2 & 98.0  & 96.9 & 95.8 & 90.1 & 94.6& 93.78 \\
                    ViTSTR-S~\cite{atienza2021vision} & Real  & 97.80& 97.90& 96.00& 96.20& 89.00& 91.50& 94.85 \\
                    DiG-ViT-B~\cite{yang2023reading} & Real  & 97.60& 97.60& 96.50& 96.50& 88.90& 92.90& 94.86 \\
                    PARSeq$_A$$^{\sharp}$~\cite{bautista2022scene} & Real  &97.32 & 97.87& 97.37& 97.22& 90.34& 94.73& 95.57\\
                    ABINet~\cite{fang2021read}& Real  &98.00& 98.60& 98.20& 97.20& 90.50& 94.10& 96.01 \\
                    MAERec-B~\cite{jiang2023revisiting} & Union14M-L  & 98.10& 98.50& 97.80& 98.60& 90.7 & 94.40& 96.15\\

    			CLIP4STR-B~\cite{zhao2023clip4str} &Real  & 98.36 & 98.73 & 97.68 & 98.96 & 91.39 & 96.74& 96.89 \\
    			CLIP4STR-L~\cite{zhao2023clip4str} &Real & 98.48  &  \textbf{99.43} & 98.15  &  98.96 & 91.66  & 97.36  &97.06 \\
       \hline
    			\bf{CLIP4STR-B*} &REBU-Syn & 99.29 & 98.96 & \textbf{98.76} & 99.65 & 92.27 & 97.83&97.25 \\
    			\bf{CLIP4STR-L*}&REBU-Syn & \textbf{99.42} &  99.13 & 98.61 & \textbf{99.65} & \textbf{92.6} & \textbf{98.13}& \textbf{97.42} \\
    			\hline	
    		\end{tabular}
            }
	\end{center}
        \setlength{\abovecaptionskip}{0.2cm}
        \setlength{\belowcaptionskip}{-0.2cm}
	\caption{Word accuracy on 6 common test benchmarks, * indicates training with REBU-Syn, Avg is the weighted average result on 6 common test benchmarks. ${\sharp}$ indicates reproduced by us.}
	\label{sota-model-table}
\end{table*}

\paragraph{Task-related pre-trained models are more effective.} The utility of pretrained models in low-level vision tasks is well-known, but their applicability in OCR tasks warrants investigation. To address this, we experimented with various pretrained models, some trained on ImageNet and others specifically for OCR recognition tasks. In the last two rows of Table ~\ref{pretrain-table}, we maintained consistent training schedules, learning rates and epochs as used for PARSeq. Intriguingly, the ImageNet-21k pre-trained models underperformed compared to those trained from scratch, a trend observed in both PARSeq and CLIP4STR models. This suggests that pretraining on non-OCR-specific tasks might not be beneficial, and can even be detrimental to OCR performance. The OCR task necessitates a connection between visual and textual elements, akin to the CLIP experiment's original purpose, whereas purely visual tasks focus more on high-level semantics and lack the textual nuances critical for OCR.

Additionally, when we trained PARSeq-S using the REBU-Syn dataset, it achieved a higher accuracy of 96.85$\%$ compared to training solely on the real data REBU. Further fine-tuning the 96.85$\%$ model with REBU led to an increased accuracy of 97.01$\%$, indicating an improvement. This demonstrates the efficacy of task-related pretrained models in OCR tasks. To attain higher accuracy, a recommended approach is \underline{training on all data first and then fine-tune on real data.}

\subsection{Comparison with SOTA Methods}

Recently, the remarkable performance of CLIP4STR across multiple benchmarks prompted us to conduct further experiments, guided by our scaling law. Initially, we focused on data composition, employing a 3:1 ratio of synthetic to real data for training the model, in conjunction with fine-tuning using a pre-trained model for related tasks. Our reproducible results led to a notable improvement in CLIP4STR-B, enhancing its accuracy from 96.54$\%$ to 97.25$\%$, an increase of 0.65$\%$. This achievement represents the best result to date in the text recognition task.

To further delve into the impact of larger models, we replicated this experiment on CLIP4STR-L. This model achieved a new state-of-the-art, recording a top-1 average accuracy of 97.42$\%$ on 6 common test benchmarks, as detailed in Table~\ref{data-ratio-table}. These findings underscore the significant role of large models in advancing the field of OCR.

%% file: sec/5_discussion.tex
\section{Discussion and Conclusion}

In this paper, we establish the presence of smooth power laws within the OCR field, demonstrating that model performance predictably improves with sufficient increases in model size, data volume and computational resources. Additionally, we identified several key principles crucial for effective model training in OCR:  1) Large-scale models more efficiently utilize samples. 2) The proportion of training data from various sources is crucial for model training; 3) Task-related pre-trained models enhance effectiveness. Beyond identifying these guiding principles, we compiled a large-scale dataset to improve the performance of the OCR model. Leveraging these rules, we successfully trained a model that achieved a new state-of-the-art average accuracy of 97.42$\%$ on the test benchmark.

We conduct extensive experiments on both model scaling and data scaling, successfully demonstrating the existence of scaling laws in OCR. Furthermore, we observe that data scaling is a particularly advantageous approach as it enhances model accuracy without incurring additional costs during training or inference. However, challenges persist in the realm of model scaling. While large-scale models exhibit superior performance with substantial data, their training is considerably more costly. Adjusting each parameter can be extremely expensive, with each training iteration potentially costing millions of dollars. To optimize the performance of these models, careful selection of the best hyperparameters during training is essential. We hope that our work can attract more researchers' attention to reduce the training cost of large-scale models.

Our experiments are based on large-scale natural scene text data sets. In the future, we will consider exploring the scaling law in more challenging text recognition data sets such as handwriting and historical texts.

%% file: sec/X_suppl.tex
\clearpage
\appendix
\label{app:proofs}
\setcounter{page}{1}
\setcounter{table}{0}   
\setcounter{figure}{0}
\maketitlesupplementary

\section{More Experiment Analysis}

\subsection{The impact of model training details}
Regarding how to train the optimal model, we conduct relevant research on multiple dimensions such as batch size and depth used during training.

\textbf{BatchSize} we focused on examining the impact of various batch sizes on the accuracy of the PARSeq-B model. This investigation was integral to determining the model's optimal training conditions. The findings, as presented in Table~\ref{batchsize-table}, reveal that the model reaches its optimal performance, with an accuracy of 96.35$\%$, when the batch size is configured to 1024. This result corroborates the conclusions from the CLIP4STR\cite{zhao2023clip4str}. it is underscoring the significant role that larger batch sizes play in enhancing model accuracy. Notably, it was also observed that an excessively large batch size leads to a reduction in accuracy, indicating a critical balance in batch size selection for optimal model training.
\begin{table}[ht!]
	\begin{center}
            \scalebox{0.7}{
    		\begin{tabular}{c|c|c|c}
    			\hline
    			{\bf{Model}} & {\bf{Backbone}}& {\bf{Batch}}& {\bf{Word Acc}} \\

    			\hline
    			PARSeq & ViT-B & 1344 & 96.28 \\
                PARSeq & ViT-B & 1024 & \textbf{96.35} \\
   			  PARSeq & ViT-B & 896 & 96.33 \\
    			PARSeq & ViT-B & 448 & 96.3 \\

    			\hline
    		\end{tabular}
            }
	\end{center}
   \setlength{\abovecaptionskip}{0cm}
   \setlength{\belowcaptionskip}{-0.3cm}
	\caption{Average accuracy using different batch size on common benchmarks, training model in real dataset.}
	\label{batchsize-table}
\end{table}

\textbf{Depth} 
PARSeq is divided into encoder and decoder. The encoder leverages the widely-recognized Vision Transformer (ViT) series, specifically employing the ViT-S variant. Conversely, the decoder is subject to more intricate fine-tuning, particularly concerning its depth. This aspect of the model architecture was a focal point of our research.

Our empirical investigations, as detailed in Table~\ref{depth-table}, centered on the interplay between the encoder's ViT-S configuration and varying depths of the decoder. The experimental findings were revealing. With the encoder consistently utilizing ViT-S, we observed that setting the decoder's depth to 1 layers resulted in optimal model accuracy. This suggests a significant relationship between decoder depth and model performance, underlining the importance of carefully calibrated model architecture in achieving high OCR accuracy. Our results contribute to a deeper understanding of the architectural nuances in Transformer-based OCR models and their impact on performance. 

\begin{table}[ht!]
	\begin{center}
            \scalebox{0.7}{
    		\begin{tabular}{c|c|c|c}
    			\hline
    			{\bf{Model}} & {\bf{Encoder}}& {\bf{Decoder-Depth}}& {\bf{Word Acc}} \\

    			\hline
    			PARSeq & ViT-S & 1 & \textbf{95.56} \\
                PARSeq & ViT-S & 2 & 95.31 \\
   			PARSeq & ViT-S & 3 & 94.77 \\
    			PARSeq & ViT-S & 4 & 94.50 \\
                PARSeq & ViT-S & 5 & 93.77 \\

    			\hline
    		\end{tabular}
            }
	\end{center}
    \setlength{\abovecaptionskip}{0cm}
    \setlength{\belowcaptionskip}{-0.3cm}
	\caption{Average accuracy using different depth for decoder on benchmark test set, training model in real dataset.}
	\label{depth-table}
\end{table}

\begin{table*}[ht]
	\begin{center}
            \scalebox{0.85}{
    		\begin{tabular}{c|c|c|c|c|c|c|c|c|c}
    			\hline
    	
    			{\bf{Method}} & {\bf{Training data}}  &  \bf{Artistic} & \bf{Contextless} & \bf{Curve}  & \bf{General} & \bf{Multi-Oriented} &  \bf{Multi-Words} & \bf{Salient} & \bf{Avg} \\ 	
    			\hline
                    CRNN ~\cite{shi2015endtoend} & MJ+ST & 20.7 & 25.6 & 7.5 & 32.0 & 0.9 & 25.6 & 13.9 & 18.0 \\
                    SVTR ~\cite{du2022svtr}& MJ+ST & 37.9 & 44.2 & 63.0 & 52.8 & 32.1 & 49.1 & 67.5 & 49.5 \\
                    MORAN ~\cite{luo2019multiobject}& MJ+ST & 29.4 & 20.7 & 8.9 & 35.2 & 0.7 & 23.8 & 17.9 & 19.5 \\
                    ASTER ~\cite{attention2} & MJ+ST & 27.7 & 33.0 & 34.0 & 39.8 & 10.2 & 27.6 & 48.2 & 31.5 \\
                    NRTR ~\cite{sheng2019nrtr} & MJ+ST & 36.6 & 37.3 & 31.7 & 48.0 & 4.4 & 54.9 & 30.6 & 34.8 \\
                    SAR ~\cite{li2019show} & MJ+ST & 42.6 & 44.2 & 44.3 & 50.5 & 7.7 & 51.2 & 44.0 & 40.6 \\
                    DAN ~\cite{wang2019decoupled} & MJ+ST & 35.0 & 40.3 & 26.7 & 42.1 & 1.5 & 42.2 & 36.5 & 32.0 \\       
                    SATRN ~\cite{lee2019recognizing} & MJ+ST & 48.0 & 45.3 & 51.1 & 58.5 & 15.8 & 52.5 & 62.7 & 47.7 \\ 
                    RobustScanner ~\cite{yue2020robustscanner} & MJ+ST & 41.2 & 42.6 & 43.6 & 39.5 & 7.9 & 46.9 & 44.9 & 38.1 \\        
                    SRN ~\cite{SRN2020Yu} & MJ+ST & 34.1 & 28.7 & 63.4 & 46.3 & 25.3 & 26.7 & 56.5 & 40.1 \\
                    ABINet ~\cite{fang2021read}& MJ+ST & 43.3 & 38.3 & 59.5 & 55.6 & 12.7 & 50.8 & 62.0 & 46.0 \\
                    VisionLAN ~\cite{wang2021one}  & MJ+ST & 47.8 & 48.0 & 57.7 & 52.1 & 14.2 & 47.9 & 64.0 & 47.4 \\
                    MATRN ~\cite{na2022multimodal} & MJ+ST & 43.8 & 41.9 & 63.1 & 57.0 & 13.4 & 53.2 & 66.4 & 48.4 \\

                    CRNN ~\cite{shi2015endtoend} &Union14M & 31.9 & 39.3 & 18.9 & 58.1 & 4.3 & 21.5 & 15.1 & 27.0 \\
                    SVTR ~\cite{du2022svtr}& Union14M  & 50.2 & 63.0 & 70.5 & 74.7 & 66.6 & 42.6 & 71.4 & 62.7 \\
                    MORAN ~\cite{luo2019multiobject}& Union14M  & 44.3 & 51.1 & 42.4 & 42.9 & 12.4 & 36.8 & 41.0 & 38.7 \\
                    ASTER ~\cite{attention2} &Union14M  & 39.2 & 47.9 & 37.4 & 64.4 & 12.5 & 34.5 & 30.2 & 38.0 \\
                    NRTR ~\cite{sheng2019nrtr} &Union14M  & 51.8 & 65.1 & 47.9 & 72.9 & 39.1 & 51.4 & 40.1 & 52.6 \\
                    SAR ~\cite{li2019show} & Union14M  & 58.0 & 69.0 & 66.9 & 73.7 & 54.7 & 51.2 & 57.0 & 61.5 \\
                    DAN ~\cite{wang2019decoupled} & Union14M  & 47.0 & 56.6 & 44.6 & 66.7 & 22.1 & 39.8 & 41.5 & 45.5 \\       
                    SATRN ~\cite{lee2019recognizing} & Union14M  & 64.3 & 71.1 & 73.0 & 78.8 & 64.7 & 47.4 & 69.2 & 66.9 \\ 
                    RobustScanner ~\cite{yue2020robustscanner} & Union14M  & 58.7 & 72.7 & 64.2 & 73.5 & 52.8 & 47.8 & 56.9 & 60.9 \\                     
                    SRN ~\cite{SRN2020Yu} & Union14M  & 47.6 & 57.9 & 48.7 & 60.7 & 20.0 & 27.9 & 41.6 & 43.5 \\
                    ABINet ~\cite{fang2021read}& Union14M  & 62.2 & 66.3 & 73.0 & 75.6 & 59.6 & 43.1 & 69.5 & 64.2 \\
                    VisionLAN ~\cite{wang2021one}  & Union14M  & 54.4 & 60.1 & 68.8 & 72.1 & 55.2 & 37.9 & 64.7 & 59.0 \\
                    MATRN ~\cite{na2022multimodal} & Union14M  & 67.3 & 71.0 & 79.3 & 78.4 & 66.0 & 53.8 & 74.9 & 70.0 \\
                    MAERec-S ~\cite{jiang2023revisiting} & Union14M-L  & 68.9 & 77.8 & 79.3 & 80.4 & 69.5 & 51.9 & 75.1 & 71.8 \\  
                    MAERec-B ~\cite{jiang2023revisiting} & Union14M-L  & 75.9 & 80.7 & 86.6 & 83.8 & 82.1 & 56.2 & 82.2 & 78.2 \\
                    
                    PARSeq-S ~\cite{bautista2022scene} & R & 81.7 & 86.5 & 91.1 & 86.5 & 89.3 & 85.3 & 84.6 & 86.5 \\
                    
                    CLIP4STR-B ~\cite{zhao2023clip4str} & R & 86.5 & \textbf{92.2} & 96.3 & 89.9 & 96.1 & 88.9 & 91.2 & 91.6\\
                    CLIP4STR-L ~\cite{zhao2023clip4str} & R& 87.2 & 91.0 & \textbf{97.0} & \textbf{90.3} & 96.6 & 89.9 & 91.5 & 91.9 \\
                    
                    \hline
    			\bf{PARSeq-S*} &REBU-Syn & 85.2 & 89.4 & 94.0 & 88.0 & 93.1 & 89.9 & 89.8 & 89.9 \\
    			\bf{CLIP4STR-B*} &REBU-Syn & 88.6 & 90.1 & 96.4 & 89.1 & 96.3 & \textbf{92.2} & 91.9 & 92.1\\
    			\bf{CLIP4STR-L*}&REBU-Syn & \textbf{88.6} & 90.4 & 96.4 & 89.3 & \textbf{97.2} & 90.7& \textbf{92.7} & \textbf{92.2} \\
    			\hline	
    		\end{tabular}
            }
	\end{center}
	\vspace{-10pt}
        \setlength{\belowcaptionskip}{-0.3cm}
	\caption{Word accuracy on Union14M benchmark, * indicates training with REBU-Syn.}
	\label{union-benchmark-table}
\end{table*}

\subsection{Benefits of pretraining in different languages}

In this supplementary section, we conduct a thorough examination of the impact of language-specific pretraining on OCR models, with a particular focus on fine-tuning for English datasets. Our approach involved utilizing models pretrained in Arabic, Latin, and a hybrid of Chinese-English, each trained on a dataset comprising 300,000 entries drawn from private sources. The core architecture for these models is based on the CMT-S framework, as detailed in Guo et al. (2022)~\cite{guo2022cmt}. Subsequent secondary training was conducted on the REB dataset, utilizing REBU-Syn, wherein different language-specific pretrained models were employed. Notably, the final classification layer's parameters were not loaded from these pretrained models to ensure a fair comparison.

As illustrated in Table~\ref{language-pretrain-table}, our results reveal pronounced improvements in models pretrained in Latin, Chinese, and English, with Latin demonstrating the most substantial enhancement. This improvement is likely due to the visual congruence between Latin and English scripts, emphasizing the OCR models' dependency on visual features for effective recognition. Meanwhile, the performance of models pretrained in Chinese and English, though slightly lower by a margin of 0.01$\%$ compared to the Latin model, indicates a potential bias introduced by the inclusion of Chinese data in the pretraining phase.

Intriguingly, models pretrained in Arabic did not exhibit significant benefits over their non-pretrained counterparts. This can be attributed to the stark visual differences between Arabic and English scripts, reinforcing the notion that visual similarity plays a crucial role in the efficacy of pretraining for OCR tasks. Collectively, these findings suggest that pretraining OCR models with languages visually akin to the target language offers enhanced benefits. Conversely, a pronounced visual dissimilarity between the scripts negates the advantages of pretraining, a critical consideration for the development of effective OCR systems.

\begin{table}[ht!]
	\begin{center}
            \scalebox{0.7}{
    		\begin{tabular}{c|c|c|c}
    			\hline
    			{\bf{Pretrain}} & \bf{Model} & {\bf{Datasets}}& {\bf{Word Acc}} \\
    			\hline
                From Scratch & PARSeq & REB  & 95.60\\
    			Arabic  & PARSeq & REB  & 95.62\\
    			Cn-En  & PARSeq & REB  & 95.81 \\
   			Latin  & PARSeq & REB  & \textbf{95.82} \\
    			\hline
    		\end{tabular}}
	\end{center}
        \setlength{\belowcaptionskip}{-0.4cm}
	\caption{Average accuracy using language-specific pretraining on benchmark test set, training model in real dataset of REB.}
	\label{language-pretrain-table}
\end{table}

\section{Comparisons on Union14M benchmark}

To evaluate the generalization capabilities of our model, we conducted an extensive assessment using the Union14M benchmark dataset~\cite{jiang2023revisiting}. This benchmark is particularly comprehensive, encompassing a vast array of real-world textual data, systematically categorized into seven distinct subsets: Artistic, Contextless, Curve, General, Multi-Oriented, Multi-Words and Salient. The results of this evaluation, presented in Table\ref{union-benchmark-table}, demonstrate the model's robust and consistent performance across a range of scenarios. Notably, In comparative evaluations against standard benchmarks and the multifaceted Union14M dataset, the CLIP4STR-L* model emerges as a standout performer. This model demonstrates exceptional accuracy across the majority of datasets. Its ability to consistently deliver high-quality results, particularly in the context of the challenging Union14M benchmark, underscores its robustness and versatility. Such performance highlights the efficacy of the CLIP4STR-L* architecture in handling a diverse range of textual data scenarios, making it a benchmark in the field. 
\begin{figure*}
  \centering
 \includegraphics[width=0.7\linewidth] 
            {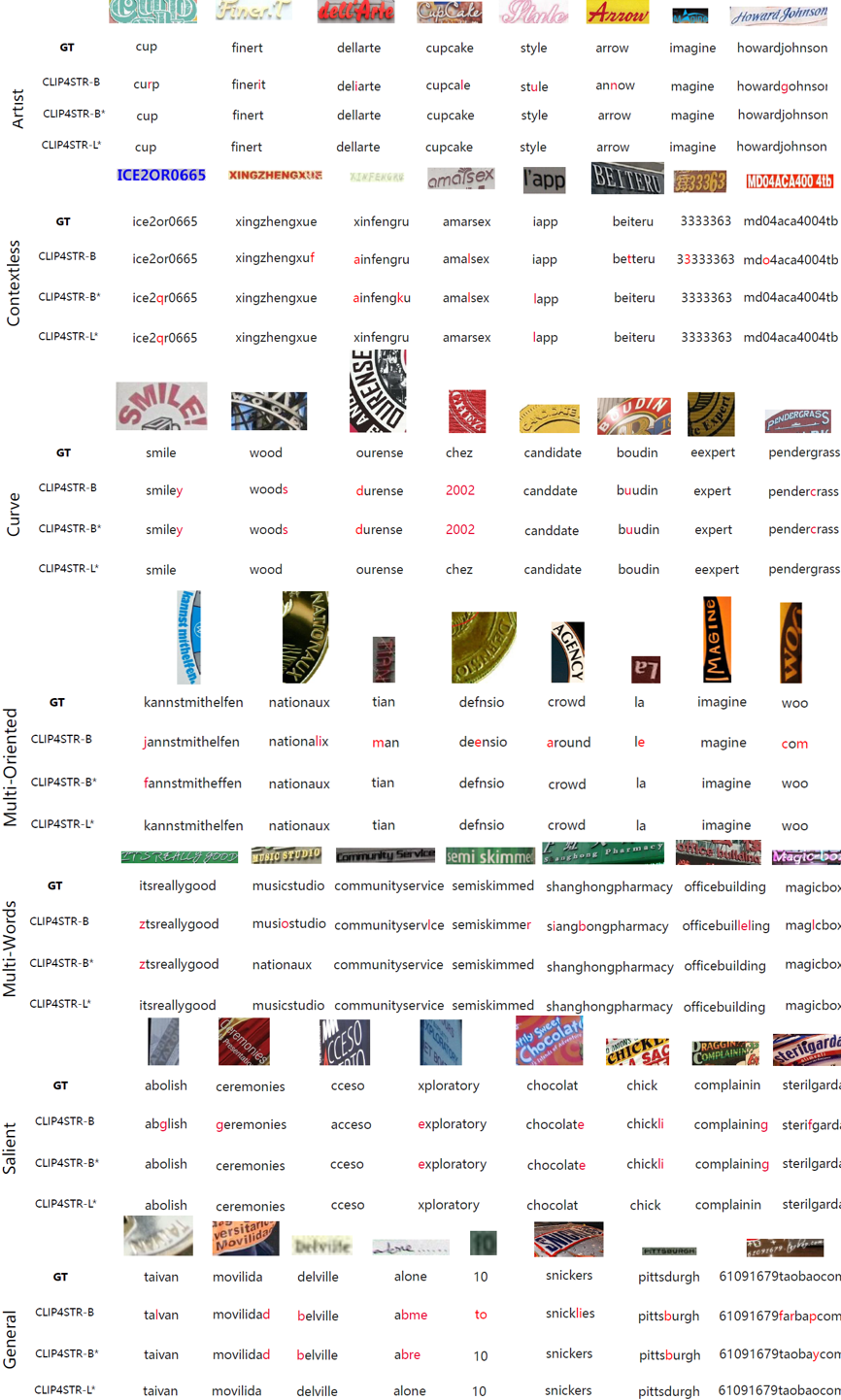} 

  \hfill

  \caption{Error analysis of the Union14M benchmark. We select three representative models and show their prediction results (Text in black
represents correct prediction and red text vice versa).}
  \label{fig:view}
\end{figure*}

\section{Visulization Analysis}
In Fig~\ref{fig:view}, we present a visualization of our model's performance across the seven major categories of the Union14M benchmark. The results demonstrate that our model outperforms in the majority of datasets. However, a slight dip in effectiveness is noted in the Contextless dataset. This can be attributed to the limitations of the text encoder in processing texts lacking semantic information.

Despite this, our model distinguishes itself from other contemporary OCR systems through its enhanced ability to accurately interpret and navigate a diverse range of complex real-world scenarios. This advancement significantly bolsters the robustness of OCR models, enabling them to operate with greater reliability in varied and challenging environments. The enhanced robustness of our model not only showcases its technical excellence but also emphasizes its practical applicability in real-world settings characterized by high variability and complexity.
\begin{figure*}
  \centering
 \includegraphics[width=0.98\linewidth] 
            {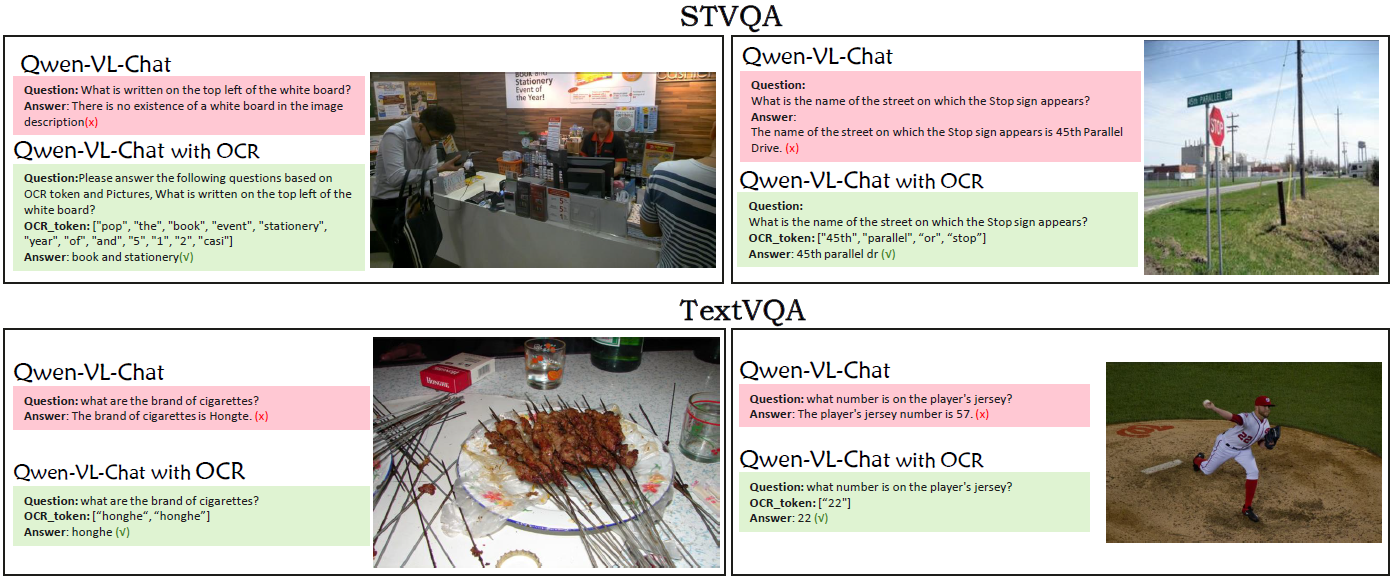} 

  \hfill
  \setlength{\abovecaptionskip}{-0.3cm}
  
  \setlength{\belowcaptionskip}{-0.5cm}
  \caption{Visual anwer comparison for QWen-VL-Chat with or without OCR in natural scenes VQA.}
  \label{fig:STVQA_TEXTVQA}
\end{figure*}
\section{OCR Enhanced LMM}
In the realm of large-scale models, we observe a distinct bifurcation into two primary categories: Large Language Models (LLMs) and Large Multimodal Models(LMMs). It is crucial to acknowledge that while LLMs are devoid of a visual component, LMMs’ visual branches demonstrate room for enhancement in terms of OCR capabilities\cite{shi2023exploring}. This observation underscores the relative underdevelopment of OCR proficiency within large-scale models. Optical Character Recognition (OCR) tasks, however, offer a promising avenue to address this shortfall, thereby motivating our investigation into the benefits of integrating OCR with these models.

 \textbf{Dataset and Metric} Our analysis utilized a diverse range of tasks from the Visual Question Answering (VQA) series, specifically STVQA~\cite{biten2019scene}, TextVQA~\cite{singh2019towards}, DocVQA~\cite{mathew2021docvqa} and InfoVQA~\cite{mathew2022infographicvqa}. While STVQA and TextVQA are geared towards natural scenes, DocVQA and InfoVQA focus on general document contexts. Here are some details of evaluation dataset:
\begin{itemize}
\item \textbf{STVQA} contains 31K questions that require understanding the scene text, based on 23K images from : ICDAR2013 and ICDAR2015, ImageNet~\cite{deng2009imagenet}, VizWiz~\cite{gurari2018vizwiz}, IIIT Scene Text Retrieval, Visual Genome~\cite{krishna2016visual} and COCO-Text. 
\item \textbf{TextVQA} contains 45K questions that need to read and reason the text in images, based on 28K images from natural images.
\item \textbf{DocVQA} contains 50K questions and 12K images from industry documents. 
\item \textbf{InfoVQA} contains 30K questions that require understanding the document text,  based on 5.4K images combining textual, graphical and visual elements from Infographics. 
\end{itemize}
We employed the Average Normalized Levenshtein Similarity (ANLS) as our evaluation metric, a standard in the VQA domain.
\begin{figure*}
  \centering
 \includegraphics[width=0.98\linewidth] 
            {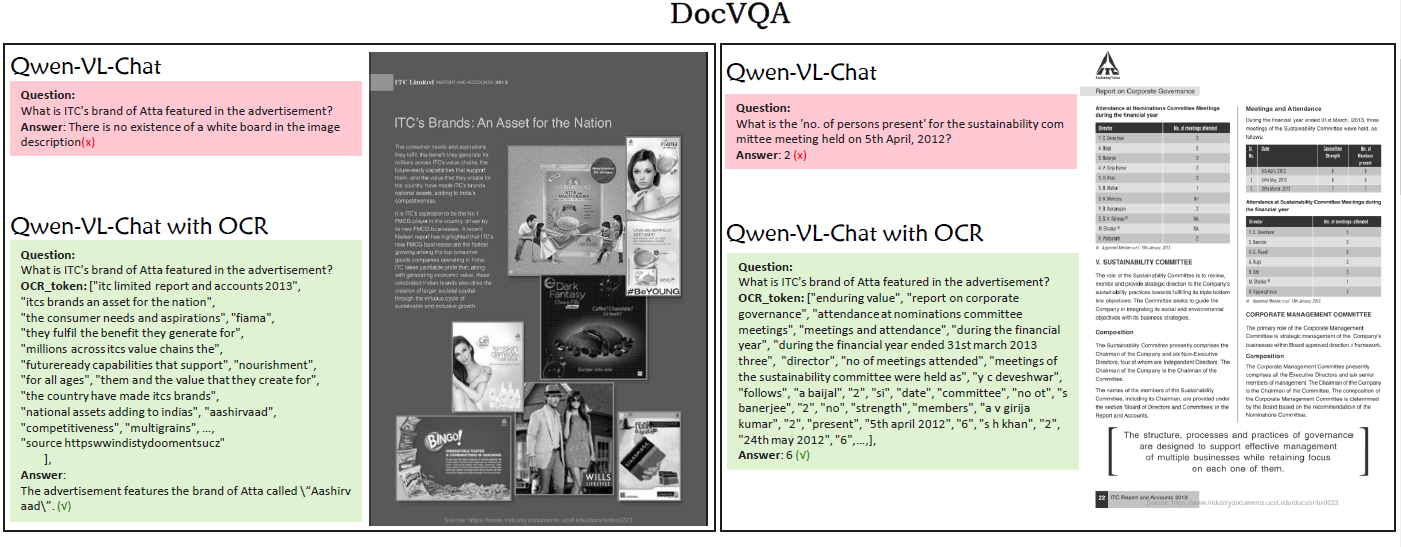} 
  \hfill
  \setlength{\belowcaptionskip}{-0.5cm}
  \caption{Visual anwer comparison for QWen-VL-Chat with or without OCR in Document VQA.}
  \label{fig:DocVQA}
\end{figure*}

\textbf{Experiment Setting} For the large-scale model, we selected the recently unveiled Qwen-VL-chat~\cite{bai2023qwen}, a state-of-the-art multimodal model. In terms of OCR, we utilized Rosetta~\cite{borisyuk2018rosetta} for detection, and CLIP4STR-L* for recognition. We began by concatenating the text recognized through coordinate information to generate OCR tokens. These tokens, combined with the question, formed our prompts. The prompt format was meticulously refined to: 'OCR token: $\left\{ocr tokens\right\}$, please answer the following questions based on OCR token and pictures, $\left\{question\right\}$'. This approach involved inputting both the prompt and images into the large-scale model.

\begin{table}[ht!]
	\begin{center}
            \scalebox{0.7}{
    		\begin{tabular}{c|c|c|c|c}
    			\hline
    			{\bf{Model}} & {\bf{STVQA}} & {\bf{TextVQA}}& {\bf{DocVQA}}&{\bf{InfoVQA}} \\
                \hline
                BLIP-2~\cite{alayrac2022flamingo} & 21.7 & 32.2 & 4.9& - \\
                LLaVAR~\cite{liu2023visual}& 39.2 & 48.5 & 11.6& - \\
                InstructBLIP~\cite{liu2023improved} & - & 50.7 & -& 38.3 \\
                
                LLaMA-7B~\cite{ye2023mplug} & - & 52.6 & 62.2 & 38.2 \\
                Pix2Struct-base~\cite{lee2023pix2struct} & - & - & 72.1 & 38.2 \\
                \hline
                Qwen-VL-Chat & 50.25 & 61.5  & 63.41 & 31.7 \\
    		 Qwen-VL-Chat with OCR  & \bf{70.32} & \bf{69.64}  & \bf{73.44} & \bf{38.48} \\
    			\hline
    		\end{tabular}}
	\end{center}
        \setlength{\abovecaptionskip}{-0.1cm}
        \setlength{\belowcaptionskip}{-0.3cm}
	\caption{Result on benchmarks for VQA tasks using LMM models with or without OCR, all result are ANLS on the val split.}
	\label{LLM-table}
\end{table}

\textbf{Result and Analyze} We performed a detailed comparative analysis to assess the accuracy of the QWen-VL-chat model, examining its performance with and without OCR integration, as delineated in Table~\ref{LLM-table}. Our results reveal a significant improvement in the accuracy of the model for scene-based VQA tasks upon the integration of OCR. Additionally, there is a noticeable enhancement in document-based VQA tasks. These findings suggest that the incorporation of OCR not only enhances the model's accuracy but also extends its generalization capabilities across diverse VQA scenarios. This evidence distinctly highlights the vital role that OCR inputs play in augmenting the performance of LVLM for downstream tasks. Furthermore, the improved accuracy with OCR integration underscores the model's enhanced ability to interpret and analyze combined visual and textual data, thereby validating the efficacy of multimodal approaches in tackling complex analytical challenges.

\textbf{VQA Visulization Analysis} Our visual analysis of QWen-VL-Chat, with and without the OCR module, across varied datasets offers critical insights. In natural scene Visual Question Answering (VQA) tasks, QWen-VL-Chat encounters difficulties in detecting small text in images. The upper left corner of Fig~\ref{fig:STVQA_TEXTVQA}, the model overlooks pertinent content, erroneously indicating its absence. Moreover, its tendency to inaccurately complete blurred text stems from its sophisticated linguistic abilities. This is evident in the upper right corner of Fig~\ref{fig:STVQA_TEXTVQA}, where 'dr' in '45th parallel dr' is incorrectly expanded to 'drive'. Notably, the model's text misidentification issues, such as converting 'honghe' to 'Hongte' on a cigarette pack as depicted in the lower left corner of Fig~\ref{fig:STVQA_TEXTVQA} (mistaking the second 'h' for a 't'), are significantly mitigated by integrating our OCR module.

In general document scenarios involving dense textual information, the performance of QWen-VL-Chat remains suboptimal.In the left of Fig~\ref{fig:DocVQA}, when tasked with identifying brands in advertisements amidst extensive text, the model erroneously generates non-existent words from the image. Incorporating OCR crucially directs the model towards accurate text recognition. This pattern is consistent in table-based VQA Tasks in the right of Fig~\ref{fig:DocVQA}, where the model frequently errs in its responses. The inclusion of OCR proves instrumental in steering the model towards correct answers. This comprehensive analysis highlights the pivotal role of OCR in augmenting LMM models' comprehension and recognition capabilities within intricate visual-textual contexts.

\section{Scaling law algorithm description}
We formalize the power law of performance in terms of scaling factors, and the implemented details are shown in Algorithm~\ref{algo_disjdecomp}.
\IncMargin{1em}
\begin{algorithm} \SetKwFunction{PowerLaw}{PowerLaw}\SetKwFunction{FitLine}{FitLine} \SetKwInOut{Input}{input}\SetKwInOut{Output}{output}
	
	\Input{x-axis data for data volume, model size or compute time $X$, word error rate $E$.} 
	\Output{$a_{0}$, $a_{1}$ are the coefficients of the power law function $E(\cdot) = (a_{0}*X)^{a_{1}}$, $v$ is used to determine whether the power law holds.}
	 \BlankLine 
      ${X}' \gets logX$, ${E}'\gets logE$;
      
      define $LineFunc({X}', {E}') = k * {X}' + b$;

        
        \For{$i\leftarrow 1$ to $t-1$}{
            \emph{Use the first t-1 points to fit the straight line equation $LineFunc({X}', {E}')$ and obtain the coefficients, $k$ and $b$.}
        }

        \emph{// Replace $({X}', {E}')$ in the straight line formula $LineFunc$ with $(X, E)$ to obtain the coefficients ($a_{0}$, $a_{1}$) of the power law function $E(\cdot) = (a_{0}*X)^{a_{1}}$.}

        $(a_{0},a_{1}) \gets$ $logE = k * logX + b$
       
        \BlankLine
        \BlankLine
        
        \emph{// Verify that ($X_t$, $E_t$) is on the equation of the power law function $E(\cdot) = (a_{0}*X)^{a_{1}}$.}
        
        $E^{pred}_{t} \gets (a_{0} * X_t)^{a_{1}} $ ; 

        $dev \gets E^{pred}_{t} - E_t$ \;
        \lIf{$dev< 0.1$} { $v \gets {1}$}
        \lElse{$v \gets 0$}
      \caption{the power-law function}
      \label{algo_disjdecomp} 
\end{algorithm}

\section{The scaling law on Union14M benchmark} 
\begin{figure}[t]
    \small
  \centering
  \begin{subfigure}{0.45\linewidth}
      \includegraphics[width=0.99\linewidth] 
            {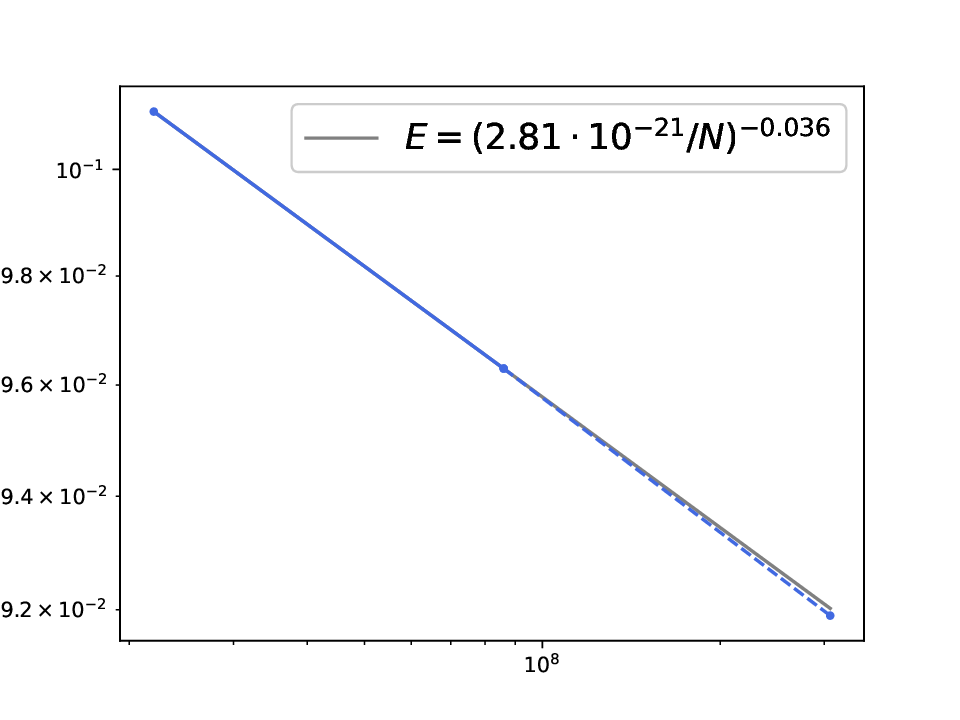} 
  \end{subfigure}
  \begin{subfigure}{0.45\linewidth}
      \includegraphics[width=0.99\linewidth] 
            {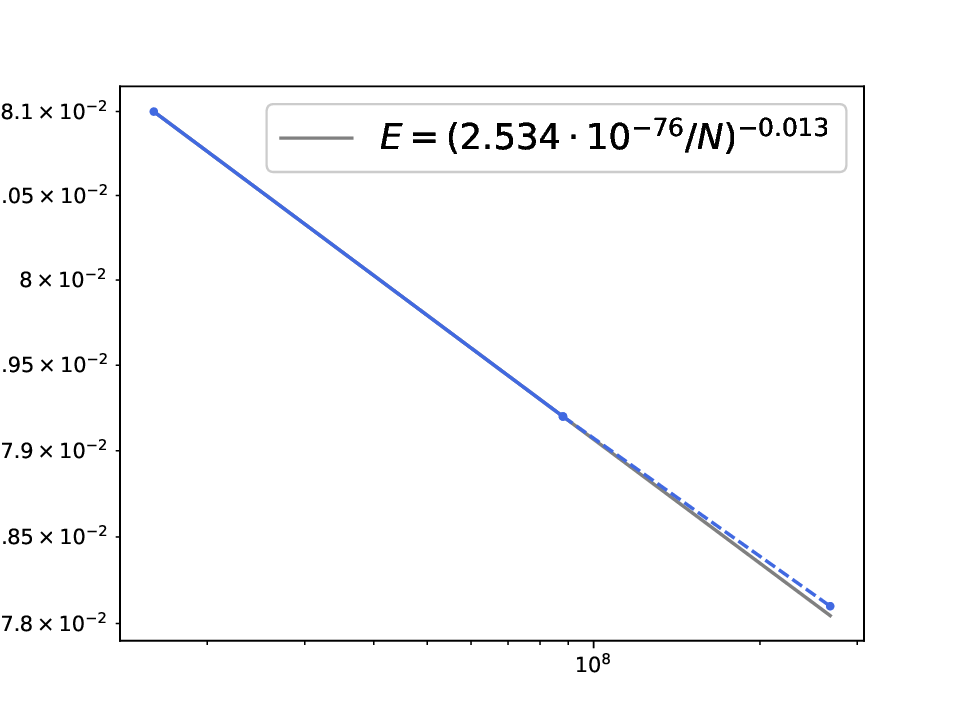} 
  \end{subfigure}
  \caption{\textbf{Left}: PARSeq-(S/B/L) results on Union14M. \textbf{Right}: CLIP4STR-(S/B/L) results on Union14M.}
  \label{fig:scale_law_for_union_clip}
\end{figure}

We supplement the experiments with scaling laws on the Union14M benchmark. The parameters and accuracy of PARSeq-(S/B/L) and CLIP4STR-(S/B/L) on the Union14M benchmark are shown in Table~\ref{clip4str-union-table} and Table~\ref{parseq-union-table} respectively. The curves of scaling law on CLIP4STR and PARSeq models are shown in Fig~\ref{fig:scale_law_for_union_clip}. It demonstrates that the scaling law is still applicable on the union14M benchmark.

\begin{table}[ht!]
	\begin{center}
            \scalebox{0.7}{
    		\begin{tabular}{c|c|c}
    			\hline
    	
    			{\bf{Method}} & {\bf{Param (M)}}  &  \bf{Avg} \\ 
    			\hline
    			
                    \bf{PARSeq-S} & 22.5 & 89.89 \\
    			\bf{PARSeq-B} & 104.0 & 90.37 \\
    			\bf{PARSeq-L} & 335.9 &  \textbf{90.81} \\
    			\hline			
    		\end{tabular}
            }
	\end{center}
	\vspace{-10pt}
       \setlength{\belowcaptionskip}{-0.5cm}
	\caption{Word accuracy with different model size of CLIP4STR. \emph{Test data}: Union14M.}
	\label{parseq-union-table}
\end{table}

\begin{table}[ht!]
	\begin{center}
            \scalebox{0.7}{
    		\begin{tabular}{c|c|c}
    			\hline
    	
    			{\bf{Method}} & {\bf{Param (M)}}  &  \bf{Avg} \\ 
    			\hline
    			
                    \bf{CLIP4STR-S} & 43.6 & 91.90 \\
    			\bf{CLIP4STR-B} & 86.7 & 92.08\\
    			\bf{CLIP4STR-L} & 268.2 &  \textbf{92.19} \\
    			\hline			
    		\end{tabular}
            }
	\end{center}
	\vspace{-10pt}
       \setlength{\belowcaptionskip}{-0.5cm}
	\caption{Word accuracy with different model size of CLIP4STR. \emph{Test data}: Union14M.}
	\label{clip4str-union-table}
\end{table}

\section{Applicability in document contexts}
We validate the power law using scaling model sizes on the document dataset in addition to the STR benchmark. The FUNSD~\cite{jaume2019funsd} dataset contains a large number of scanned documents, and each sample is annotated with detailed text, word bounding boxes, and structured tags. It is intended to support the development and assessment of model performance by researchers for the purpose of processing and comprehending information from scanned documents in noisy, real-world.
Notably, CLIP4STR-L* achieved a SOTA accuracy of 96.5$\%$, surpassing the previous best, CLIP4STR-L. The experimental results are shown in Table~\ref{clip4str-funsd-table}. These results highlight the robustness of CLIP4STR-L* in both scene and document text recognition tasks.
\begin{table}[ht!]
	\begin{center}
            \scalebox{0.7}{
    		\begin{tabular}{c|c}
    			\hline
    			{\bf{Model}} &  {\bf{Word Acc}} \\

    			\hline
    			CLIP4STR-L & 96.02 \\
                CLIP4STR-L* & \textbf{96.50} \\
    			\hline
    		\end{tabular}
            }
	\end{center}
    \setlength{\abovecaptionskip}{0cm}
    \setlength{\belowcaptionskip}{-0.3cm}
	\caption{Accuracy for CLIP4STR-L on FUNSD.}
	\label{clip4str-funsd-table}
\end{table}


